\documentclass[conference]{IEEEtran}
\IEEEoverridecommandlockouts
\usepackage{amsmath,amsfonts}
\usepackage{array}
\usepackage[caption=false,font=normalsize,labelfont=sf,textfont=sf]{subfig}
\usepackage{textcomp}
\usepackage{stfloats}
\usepackage{url}
\usepackage{verbatim}
\usepackage{graphicx}
\usepackage{times}
\usepackage{soul}
\usepackage[hidelinks]{hyperref}
\usepackage[utf8]{inputenc}
\usepackage[small]{caption}
\usepackage{graphicx}
\usepackage{algpseudocode}
\usepackage{algorithm}
\usepackage{amssymb}
\usepackage{mathrsfs}
\usepackage{booktabs}
\usepackage[switch]{lineno}
\usepackage{orcidlink}
\hypersetup{hidelinks}
\def\BibTeX{{\rm B\kern-.05em{\sc i\kern-.025em b}\kern-.08em
    T\kern-.1667em\lower.7ex\hbox{E}\kern-.125emX}}
\begin{document}

\title{CuDA2: An approach for Incorporating Traitor Agents into Cooperative Multi-Agent Systems
}
\author{Zhen Chen, 
Yong Liao,
Youpeng Zhao,
Zipeng Dai,  
Jian Zhao
}
\maketitle

\begin{abstract}
Cooperative Multi-Agent Reinforcement Learning (CMARL) strategies are well known to be vulnerable to adversarial perturbations.
Previous works on adversarial attacks have primarily focused on white-box attacks that directly perturb the states or actions of victim agents, often in scenarios with a limited number of attacks.
However, gaining complete access to victim agents in real-world environments is exceedingly difficult.
To create more realistic adversarial attacks, we introduce a novel method that involves injecting traitor agents into the CMARL system.
We model this problem as a Traitor Markov Decision Process (TMDP), where traitors cannot directly attack the victim agents but can influence their formation or positioning through collisions.
In TMDP, traitors are trained using the same MARL algorithm as the victim agents, with their reward function set as the negative of the victim agents' reward. 
Despite this, the training efficiency for traitors remains low because it is challenging for them to directly associate their actions with the victim agents' rewards.
To address this issue, we propose the Curiosity-Driven Adversarial Attack (CuDA2) framework.
CuDA2 enhances the efficiency and aggressiveness of attacks on the specified victim agents' policies while maintaining the optimal policy invariance of the traitors.
Specifically, we employ a pre-trained Random Network Distillation (RND) module, where the extra reward generated by the RND module encourages traitors to explore states unencountered by the victim agents.
Extensive experiments on various scenarios from SMAC demonstrate that our CuDA2 framework offers comparable or superior adversarial attack capabilities compared to other baselines.

\end{abstract}

\section{Introduction}
Cooperative Multi-Agent Reinforcement Learning (CMARL) has recently garnered significant attention~\cite{gronauer2022multi,zhao2023mcmarl,zhao2022ctds}, finding applications in diverse areas such as autonomous vehicle teams~\cite{peng2021learning}, multi-agent pathfinding~\cite{greshler2021cooperative}, multi-UAV control~\cite{yun2022cooperative}, and dynamic algorithm configuration~\cite{xue2022multi}.
Existing CMARL methods primarily address challenges like non-stationarity~\cite{papoudakis2019dealing}, credit assignment~\cite{wang2021towards}, and scalability, all aimed at enhancing coordination in complex scenarios~\cite{christianos2021scaling}.
Both value-based methods~\cite{sunehag2017value,rashid2020monotonic} and policy gradient-based methods~\cite{foerster2018counterfactual,yu2022surprising} have shown significant coordination capabilities across a variety of tasks, such as SMAC~\cite{samvelyan19smac} and Hanabi~\cite{guo2022towards}.

While CMARL has demonstrated remarkable success across various domains, it shares a vulnerability with Single-Agent Reinforcement Learning (SARL)~\cite{zhao2022improving} to adversarial attacks. 
For example, studies such as ~\cite{yuan2023robust,sun2020stealthy} explore implementing attacks with a limited number of attempts by injecting adversarial samples at critical moments to cause the most severe damage to the agent.
Research by ~\cite{gleave2019adversarial,guo2021adversarial} investigates poisoning attacks on multi-agent reinforcement learners, assuming the attacker controls one of the learners, typically an opponent of the victim agents.
However, these attack methods necessitate full access to and control over the environment or agents, requiring advanced hacking skills to modify the server or the real/simulated environment.
In light of these limitations, we propose a more practical attack method: incorporating traitor agents into cooperative multi-agent systems.
For instance, in a soccer game, an agent could be introduced that deliberately plays poorly, or in a mobile base station environment, a base station could be introduced to interfere with the connection signals of other base stations.
This approach does not require modifying or directly operating the environment or the victim agents, making it a more feasible and realistic adversarial strategy.



In response to this type of attack, we design a CMARL scenario involving traitors to indirectly target victim agents.
Specifically, we model the problem as a Traitor Markov Decision Process (TMDP), where traitors and victim agents are on the same team but have opposing objectives.
In this setup, traitors cannot directly attack the victim agents but can influence their observations by maneuvering or colliding with them.
The success of adversarial policies stems from the ability of an attacker to manipulate the victim agents' observations by taking unconventional actions, leading the game into unfamiliar states.
This often causes the victim agents to exhibit undesired, sub-optimal behaviors.
Thus, the effectiveness of an attack may be significantly influenced by the attacker's capability to explore such states and exploit these vulnerabilities.

This insight motivates us to propose a Curiosity-Driven Adversarial Attack (CuDA2) framework, which employs a Random Network Distillation (RND) module to characterize the novelty of the victim agents' states.
Specifically, we first pre-train an RND module in an environment where traitors take random actions.
This pre-training aims to provide an intrinsic reward through the RND module when the traitors' actions cause significant displacement of the victim agents, thereby guiding the traitors to more effectively attack the victim agents.
To address the issue that additional rewards might alter the traitors' optimal policy, we use the RND module as a potential function and apply the dynamic potential-based reward shaping method to generate intrinsic rewards during the traitors' training.
We theoretically prove that this combination can ensure the invariance of the traitors' optimal policy.

To evaluate the proposed method, we conducted extensive experiments on multiple SMAC maps with varying numbers of traitors and compared CuDA2 with several baselines. 
Empirical results demonstrate that the CuDA2 framework significantly enhances the attack and disruption capabilities of the traitors.
Additionally, we performed ablation studies and visualization experiments. 
The results indicate that our method more effectively reduces the win rate of the victim agents and achieves curiosity-driven adversarial attacks more efficiently compared to algorithms that solely use the RND module.
We provide the CMARL community with a new, more practical attack method, and defending against this type of attack can enhance the robustness and security of CMARL.

\section{Related Work}
\subsection{Multi-Agent Reinforcement Learning (MARL)}
In recent year, there exist significant research progress~\cite{gronauer2022multi, hernandez2019survey} in MARL. 
Numerous methods have emerged as effective strategies for promoting coordination among agents, which can generally be categorized into policy-based and value-based methods.
Examples of policy gradient-based methods that focus on optimizing multi-agent policies include MADDPG~\cite{lowe2017multi}, COMA~\cite{foerster2018counterfactual}, DOP~\cite{wang2020dop}, and MAPPO~\cite{guo2022towards}.
MADDPG utilizes the CTDE (Centralized Training Decentralized Execution) paradigm to train policies and refine them via DDPG~\cite{lillicrap2015continuous}.
COMA also uses a centralized critic for policy optimization but incorporates a counterfactual model to determine each agent's marginal contribution in a multi-agent system.
DOP advances this approach by employing a centralized linear mixing network to break down global rewards in a cooperative system, significantly enhancing the performance of MADDPG and COMA.
Recently, MAPPO has applied the widely validated proximal policy optimization technique from single-agent reinforcement learning to the MARL domain.

Another branch of MARL methods, called value-based approaches, primarily concentrates on the factorization of the value function.
VDN~\cite{sunehag2017value} aims to break down the team value function into individual agent values using a simple additive factorization.
Adhering to the Individual-Global-Max (IGM) principle~\cite{son2019qtran}, QMIX~\cite{rashid2020monotonic} enhances value function decomposition by employing a non-linear mixing network to approximate a monotonic function value decomposition.
QPLEX~\cite{wang2020qplex} utilizes a duplex dueling network architecture to factorize the joint value function, fully exploiting the expressive power of IGM.
Research by~\cite{wang2021towards} conducted a theoretical analysis of IGM by applying a multi-agent fitted Q-iteration algorithm.
This paper primarily uses QMIX, MAPPO and VDN as the main algorithms for the experiments.

\subsection{Adversarial Attacks on MARL}
Adversarial attacks involve the intentional manipulation of machine learning models by attackers using specially crafted input samples to deceive or mislead the models.
As deep learning rapidly advances, attackers are continuously developing new attack methods, such as poisoning attacks~\cite{chen2023tutorial}, adversarial machine learning~\cite{huang2011adversarial}, and other technologies~\cite{song2019membership,wenger2021backdoor}, making it increasingly challenging to detect and defend against these attacks.
Poisoning attacks typically occur during the training phase, compromising performance and reliability through harmful data. 
Similarly, in Deep Reinforcement Learning (DRL), attackers manipulate input data during training, introducing prediction biases into the model. 
The goal of adversarial machine learning is to enhance robustness and security by examining potential attacks and threats. 
These approaches are fundamentally similar to attacks on DRL. 
This paper specifically examines the vulnerability of the DRL model, primarily through adversarial machine learning methods.

Adversarial attacks in DRL can be categorized into reward-based attacks~\cite{zhang2020adaptive}, strategy-based attacks~\cite{vahid2019adversarial,mo2022attacking}, observation-based attacks~\cite{hussenot2019copycat,li2022deep,sun2020stealthy}, environment-based attacks~\cite{bai2018adversarial,chen2018gradient}, and action-based attacks~\cite{lee2020spatiotemporally} according to their algorithmic principles. 
Reward-based attacks involve altering the reward signal from the environment, either by changing the reward value's sign or replacing the original reward function with an adversarial one. 
Strategy-based attacks use adversarial agents to generate states and actions beyond the victim agent's comprehension, causing disarray. 
Observation-based attacks involve adding perturbations to the observed image, compelling the victim agent to take actions desired by the attacker, typically by perturbing the agent's image sensor. 
Environment-based attacks modify the agent's training environment directly, altering the dynamic model or adding obstacles. 
Action-based attacks directly modify action outputs by changing the action space in the training data. 

In previous works on observation-based and action-based attacks, attackers added appropriate perturbations to observation images or actions over a period of time to mislead the victim agent into making incorrect decisions~\cite{li2023ats,yuan2023robust}. 
While ensuring the stealth of the attack, they aimed to minimize the cumulative reward and reduce the overall team's gains. 
Current research mainly focuses on achieving more efficient attacks under limited attack opportunities. 
However, these works have an important premise: the ability to obtain all permissions of the victim agent, including control over state actions. 
This means their attack methods are white-box attacks, which are known to be difficult to implement in real-world scenarios. 
Therefore, in this work, we adopt traitor setting to attack cooperative multi-agent scenarios. 
By training traitor agents, we aim to minimize the cumulative reward of the team.

\section{Background}
\subsection{Markov Decision Process (MDP)}
\label{prelim}
A standard Markov Decision Process (MDP) can be defined as a tuple ($\mathcal{S}$, $\mathcal{A}$, $\mathcal{P}$, $\mathcal{R}$, $\gamma$) where $s \in \mathcal{S}$ is the state space, $a \in \mathcal{A}$ is the action space, $\mathcal{P}(s' \mid s,a):\mathcal{S} \times \mathcal{A} \times \mathcal{S} \rightarrow [0,1]$ is the transition probability, $\mathcal{R}(s,a):\mathcal{S} \times \mathcal{A} \rightarrow \mathbb{R}$ is the reward function and $\gamma \in [0,1)$ is the discount factor, which represents the preference for immediate reward over long-term reward. 
The agent’s goal is to find the policy $\pi^*$ which at any given $s_n \in \mathcal{S}$ maximizes the expected discounted sum of rewards, 
\begin{align}
\label{eq1}
\pi^* = \mathop{argmax}\limits_{\pi}\mathbb{E}_{\pi}\left[\sum\limits_{n=0}^{K}\gamma^nR(s_n,a_n)\right],
\end{align}
where $K$ is the number of time steps in each episode. 
$K$ can be either finite or infinite, depending on whether we are using an environment with finite or infinite horizons.
\begin{align}  
\label{eq2}
Q(s,a)&=\mathbb{E}_{\pi}\left[\sum\limits_{n=0}^{K}\gamma^nR(s_n,a_n) \mid s_0=s,a_0=a\right].
\end{align}
The Q-function $Q(s,a)$ estimates how good it is to perform an action in a state~\cite{bellman1957markovian}, given the policy $\pi$.
    
\subsection{Potential Based Reward Shaping}
Reward shaping involves enhancing the original reward function by incorporating domain-specific knowledge.
This process typically uses an additive form of reward shaping. 
Formally, it can be expressed as $r' = r + F$, where $r$ represents the original reward function, $F$ is the shaping reward function, and $r'$ denotes the modified reward function. 
Early studies on reward shaping~\cite{dorigo1994robot,randlov1998learning} focused on the design of the shaping reward function $F$ but overlooked the possibility that these shaping rewards might alter the optimal policy. 
In other words, reward shaping might lead to the phenomenon of reward hacking, causing the agent to develop suboptimal strategies. 
Although~\cite{carta2022eager,zhang2021noveld} have attempted to mitigate this issue by avoiding repeated extra rewards, these approaches tend to address the symptoms rather than the root cause.

Potential-Based Reward Shaping (PBRS)~\cite{ng1999policy} was the first method to ensure the policy invariance property. Specifically, PBRS defines $F$ as the difference between potential values: 
\begin{align}
\label{eq3}
F(s_n,s_{n+1}) = \gamma\Phi(s_{n+1})-\Phi(s_n),
\end{align}
where $\Phi(s):\mathcal{S}\to \mathbb{R}$ is a potential function that provides insights into the states. Notable variations of PBRS include the potential-based advice approach: 
\begin{align}
\label{eq4}
F(s_n,a_n,s_{n+1},a_{n+1}) = \gamma\Phi(s_{n+1},a_{n+1})-\Phi(s_n,a_n),
\end{align}
which extends $\Phi$ over the state-action space for action advice~\cite{wiewiora2003principled}, the dynamic PBRS approach: 
\begin{align}
\label{eq5}
F(s_n,t_n,s_{n+1},t_{n+1}) = \gamma \Phi(s_{n+1},t_{n+1})-\Phi(s_n,t_n),
\end{align}
which incorporates a time parameter into $\Phi$ to allow for dynamic potentials~\cite{devlin2012dynamic}, and the dynamic potential-based advice approach that learns an auxiliary value function to transform any rewards into potentials~\cite{harutyunyan2015expressing}.


\begin{figure*}
    \centering
    \includegraphics[width=0.99\linewidth]{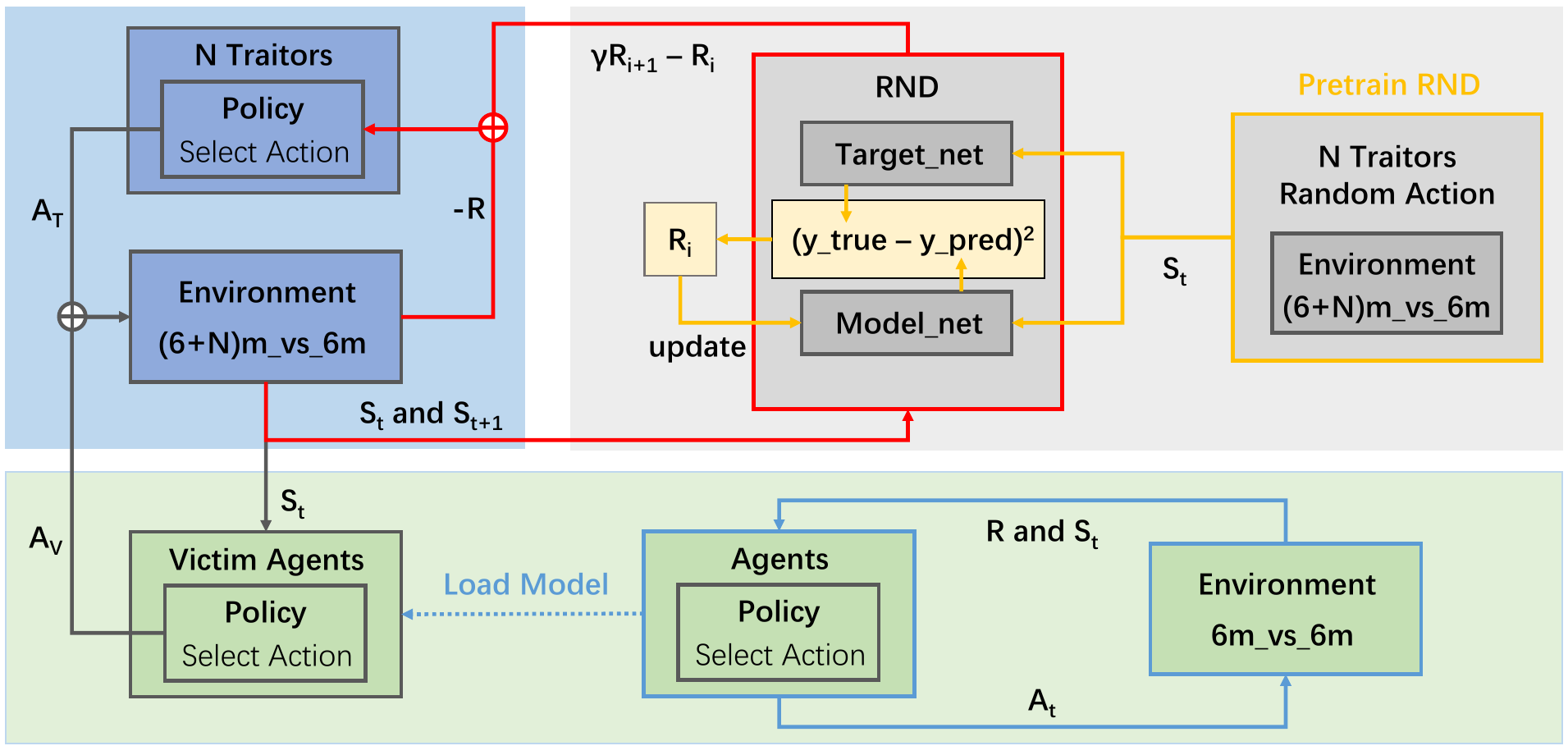}
    \caption{CuDA2 framework. First, as indicated by the green box in the figure, we need to define the target that the traitors intend to attack: pre-training and saving a model of the victim agents. Second, as shown by the gray box in the figure, we also need to pre-train the RND module within the strategy where the traitors take random actions. This can reduce the prediction error caused by the state changes of the victim agents resulting from the traitors' random actions. Finally, before training the traitors, we will load the victim agents model. During the training process, we use the pre-trained RND module as a potential function to provide the traitors with intrinsic rewards through the dynamic PBRS method.}
    \label{fig:framework}
\end{figure*}

\section{Method}
Different from previous work based on observation and action attacks, this paper introduces traitors into the training framework of the existing agents. 
The traitors belong to the victim agents' side but aims to minimize the team's win rate. 
Under this setting, as shown in Fig.\ref{fig:framework}, we propose the Curiosity-Driven Adversarial Attacks (CuDA2) framework to train the traitors. 
By pre-training the Random Network Distillation (RND) module~\cite{burda2018exploration}, it can efficiently conduct adversarial attacks on the victim agents. 
Combining RND with the dynamic PBRS method, we also theoretically prove that curiosity-driven adversarial attacks within the CuDA2 framework do not alter the traitors' optimal policy.
\subsection{Traitors Optimization Objective}

We extend the MDP model to include an action selection function for traitors $ \mathcal{P_T}(a \mid \pi_T, s) $. 

\bigskip

\textbf{Definition 1.} A \textit{Traitor Markov Decision Process} (TMDP) is a tuple 
$ \tilde{\mathcal{M}} = (\mathcal{S}, \mathcal{A}, \mathcal{P_V}, \mathcal{P_T}, \mathcal{R}, \gamma) $
where $ \mathcal{M} = (\mathcal{S}, \mathcal{A}, \mathcal{P_V}, \mathcal{R}, \gamma) $ is an MDP and the transition probability for victim agents is the special case in which the policy is applied without modification: $\mathcal{P_V}(a \mid \pi_V, s) = \pi_V(a \mid s) $.
$\mathcal{P_T}(a \mid \pi_T, s) = \Pr(A_n = a \mid \Pi = \pi_T, S_n = s) $
is the probability that action $a$ is selected in state $s$ given a policy $\pi$.
We also write $ \tilde{\mathcal{M}} = (\mathcal{M}, P_A). $

\bigskip

In the TMDP, $\pi_V$ is a fixed, non-updating policy of the victim agents. 
$R(s_n,a_n)$ is the reward obtained by the victim agents in the environment. 
The traitors' objective is to minimize the victim agents' reward, so its reward function is $R_T = -R(s_n,a_n)$. 
An optimal traitor policy for a TMDP is one that maximizes the expected return:
\begin{align}
\label{eq6}
\pi_T^* = \mathop{argmax}\limits_{\pi}\mathbb{E}_{\pi}\left[\sum\limits_{n=0}^{K}-R(s_n,a_n)\gamma^n\right],
\end{align}
where actions are sampled according to $\mathcal{P_T}(a \mid \pi_T, s)$.
\subsection{Pre-training}
Before starting to train the traitors' policy $\pi_T$, we first need to pre-train a victim agents' policy $\pi_V$ to serve as our attack target. 
Additionally, to achieve more effective attacks, we also need to pre-train an RND module.
\subsubsection{Victim Agents}
As shown in Fig.\ref{fig:framework}, the green box represents the pre-training process of the victim agents. 
The environment used is 6m-vs-6m or 8m-vs-8m, where the number of allied agents and enemies is equal, and no traitor is introduced yet. 
After a period of training, we will obtain a well-trained policy $\pi_V$ which can defeat all the enemies while minimizing its own losses and then we will save its network parameters.
\begin{align}
\label{eq7}
\pi_V^* = \mathop{argmax}\limits_{\pi}\mathbb{E}_{\pi}\left[\sum\limits_{n=0}^{K}R(s_n,a_n)\gamma^n\right].
\end{align}
 The environment used is (6+N)m-vs-6m or (8+N)m-vs-8m, where N is the number of traitors. During the traitors' training, we will load this policy, and the victim agents will generate actions based on it. The traitors' attack objective is to disrupt the victim agents' policy, making it ineffective against the enemies and thus reducing its win rate.
\subsubsection{RND module}
RND is a method used to measure the novelty of states encountered by an agent during training~\cite{xu2021deep}. 
The core idea involves two neural networks: a target network and a predictor network. 
The target network is randomly initialized at the beginning and remains fixed. 
It maps an observation from the environment to an embedding space, represented mathematically as $f : \mathcal{O} \rightarrow \mathbb{R}^k$, where $\mathcal{O}$ denotes the set of observations and $\mathbb{R}^k$ is the $k$-dimensional embedding space. 
The predictor network, denoted by $\hat{f}$, is trained to approximate the output of the target network.
It is parameterized by $\theta_{\hat{f}}$ and maps observations to the same embedding space: $\hat{f} : \mathcal{O} \rightarrow \mathbb{R}^k$. 
The predictor network is trained using gradient descent to minimize the mean squared error (MSE) between its output and the target network's output, formulated as: 
\begin{align}
\label{eq8}
\text{MSE} = \mathbb{E}[\lVert \hat{f}(x; \theta) - f(x) \rVert^2],
\end{align}
where $x$ represents the observations. 

The essence of RND is that the predictor network will perform poorly on novel states (states it has not encountered before) because it has not had the chance to learn these states during training. 
This results in a higher prediction error for novel states compared to familiar states. 
Consequently, the prediction error from the RND can be used as an intrinsic reward signal to encourage the agent to explore new and unseen areas of the state space. 
As shown in Fig.\ref{fig:framework}, the yellow line within the gray box represents the RND pre-training process.

\renewcommand{\thealgorithm}{CuDA2} 
    \begin{algorithm}
        \caption{}
        \label{alg:pseudocode}
        \begin{algorithmic}[0] 
            \Require  number of steps for each episode $K$,  the policy of victim agents $\pi_V$, the policy of traitors $\pi_T$, the intrinsic reward from random network distillation module $RND$, replay buffer $\mathcal{B}$
            \State $ \verb|\\|$\textit{pre-train the RND module}
            \For{each episode}
                \For{$n \gets 0 \to K$}
                    \State sample $a_n \sim Uniform(a_n)$
                    \State sample $s_{n+1} \sim p(s_{n+1} \mid s_n,a_n)$
                    \State update the model network of $RND$ using $s_{n}$
                \EndFor
            \EndFor
            \State $ \verb|\\|$\textit{train the policy of traitors}
            \For{each episode}
                \For{$n \gets 0 \to K$} 
                    \State sample $a_V \sim \pi_V(a_V \mid s_n)$
                    \State sample $a_T \sim \pi_T(a_T \mid s_n)$
                    \State $s_{n+1},r_V,done \gets env.step([a_V, a_T])$
                    \If{Not done}
                        \State $r_{shape} \gets \gamma RND(s_{n+1},t_{n+1}) - RND(s_n,t_n)$
                    \Else
                        \State $r_{shape} \gets 0 - RND(s_n,t_n)$
                    \EndIf
                    \State $r_T \gets -r_V + r_{shape}$
                    \State store transition $(s_n,a_T,s_{n+1},r_T)$ in $\mathcal{B}$
                    \State update the model network of $RND$ using $s_{n}$
                    \State sample random minibatch of transitions from $\mathcal{B}$
                    \State update the policy of traitors $\pi_T$
                \EndFor
            \EndFor
        \end{algorithmic}
    \end{algorithm}

\subsection{CuDA2 Framework}
As shown in Algorithm \ref{alg:pseudocode} and Fig.\ref{fig:framework}, 
the process of our framework begins with pre-training the policy of victim agents and the RND module.
During the traitors' training phase, each episode starts by resetting the environment to an initial state. 
At each time step, actions $a_V$ and $a_T$ are sampled separately for victim agents and traitors from their respective policies $\pi_V$ and $\pi_T$. 
These actions are combined and executed, resulting in a new state and a reward for the victim agents. 
The intrinsic reward for traitors is calculated based on state novelty changes determined by the RND module. 
Specifically, we get the corresponding outputs $R_{i+1}$ and $R_i$ by inputting the current state and the next state to the RND module and compute $\gamma R_{i+1} - R_i$ as intrinsic reward. 
The traitors' reward is shaped by adding the intrinsic reward to the negative victim's reward, promoting the exploration of states detrimental to the victims. 
To be noted, classic RND techniques typically adopt $R_i$ directly to encourage exploration, which may lead to reward hacking and impede the learning of an optimal policy.
In contrast, we utilize a dynamic PBRS method to ensure policy invariance, thus helping the traitor agents to grasp optimal policy.

The transition $(s_n,a_T,s_{n+1},r_T)$(current state, traitors' action, next state, traitors' reward) is stored in a replay buffer, and random samples from this buffer are used to periodically update the traitors' policy, optimizing their actions to minimize the victim agents' rewards. 
At the end of each time step, we will update the RND model with current state $s_{n}$. 
This approach enables traitors to learn strategies that maximize the impact and disruption of victim agents through intrinsic reward-driven exploration and exploitation without changing the optimal policy.

\subsection{Optimal Policy Invariance Theory Analysis}
To prove that the reward shaping of the CuDA2 framework can maintain the optimal policy, let us consider the return $U_i$ for any arbitrary agent $i$ when experiencing sequence $\bar{s}$ in a discounted framework without shaping. Formally:
\begin{align}
\label{eq9}
U_{i}(\bar{s}) = \sum_{j=n}^{K} \gamma^{j-n} r_{j,i},
\end{align}
where $r_{j,i}$ is the reward received at time $j$ by agent $i$ from the environment. Given this definition of return, the true $Q$-values can be defined formally by:
\begin{align}
\label{eq10}
Q_i(s_n, a_n) = \sum_{\bar{s}} Pr(\bar{s} | s_n, a_n) U_i(\bar{s}).
\end{align}
According to Eq.(\ref{eq5}), we now consider the same agent but with a reward function modified by adding a dynamic potential-based reward function of the form given below:
\begin{align}
\label{eq11}
F(s_n,t_n,s_{n+1},t_{n+1}) =&\gamma RND(s_{n+1},t_{n+1})\nonumber\\
&-RND(s_n,t_n),
\end{align}
where the RND function is:
\begin{align}
\label{eq12}
RND(s_n,t_n) = \lVert \hat{f}(s_n; \theta_{t_n}) - f(s_n) \rVert^2.
\end{align}
The shaped reward function $r'$ is:
\begin{align}
\label{eq13}
r'_{j,i} = r_{j,i} + F(s_j,t_j,s_{j+1},t_{j+1}).
\end{align}
The return of the shaped agent $U_{i, F}$ experiencing the same sequence $\bar{s}$ is:
\begin{align}
\label{eq14}
U_{i, F}(\bar{s}) =&\sum_{j=n}^{K} \gamma^{j-n} r'_{j,i}\nonumber\\
=& \sum_{j=n}^{K} \gamma^{j-n} (r_{j,i} + F(s_j, t_j, s_{j+1}, t_{j+1}))\nonumber\\
=& \sum_{j=n}^{K} \gamma^{j-n} (r_{j,i} + \gamma RND(s_{j+1}, t_{j+1}) - RND(s_j, t_j))\nonumber\\
=& \sum_{j=n}^{K} \gamma^{j-n} r_{j,i} + \sum_{j=n}^{K} \gamma^{j-n+1} RND(s_{j+1}, t_{j+1})\nonumber\\
&- \sum_{j=n}^{K} \gamma^{j-n} RND(s_j, t_j)\nonumber\\
=& U_i(\bar{s}) - RND(s_n, t_n) + \gamma^{K-n+1}RND(s_{K+1},t_{K+1}) \nonumber\\
=& U_i(\bar{s}) - \lVert \hat{f}(s_n; \theta_{t_n}) - f(s_n) \rVert^2 .
\end{align}
In the above equation, it is important to note that if $K$ approaches infinity, then $\gamma^{K-n+1}RND(s_{K+1},t_{K+1})$ can be ignored.
However, if this is an episodic reinforcement learning task (game, etc), where $K$ is finite and $s_{K+1}$ is a terminal state, the output of RND module ($RND(s_{K+1},t_{K+1})$) needs to be replaced with 0.
Otherwise, the potential function of the terminal state will affect the policy learned by the agent~\cite{grzes2017reward}.

By combining Eq.(\ref{eq10}) and Eq.(\ref{eq14}) we know the shaped Q-function is:
\begin{align}
\label{eq15}
Q_{i}^*(s_n, a_n) =& \sum_{\bar{s}} Pr(\bar{s} | s_n, a_n) U_{i,F}(\bar{s})\nonumber\\
=& \sum_{\bar{s}} Pr(\bar{s} | s_n, a_n) (U_i(\bar{s}) - \lVert \hat{f}(s_n; \theta_{t_n}) - f(s_n) \rVert^2)\nonumber\\
=& \sum_{\bar{s}} Pr(\bar{s} | s_n, a_n) U_i(\bar{s}) \nonumber\\
&- \sum_{\bar{s}} Pr(\bar{s} | s_n, a_n) \lVert \hat{f}(s_n; \theta_{t_n}) - f(s_n) \rVert^2\nonumber\\
=& Q_i(s_n, a_n) - \lVert \hat{f}(s_n; \theta_{t_n}) - f(s_n) \rVert^2 ,
\end{align}
where $t_n$ is the current time.

Therefore, any policy that optimizes $Q_{i}(s_n,a_n)$ also optimizes $Q_i^*(s_n,a_n)$. Since $\lVert \hat{f}(s_n; \theta_{t_n}) - f(s_n) \rVert^2$ does not depend on the action chosen, which means the choice of the optimal action in the current state is not affected by the value of this extra function. 
The reward shaped by the CuDA2 framework will not change the optimal policy.

\begin{figure}
    \centering
    \includegraphics[width=0.99\linewidth]{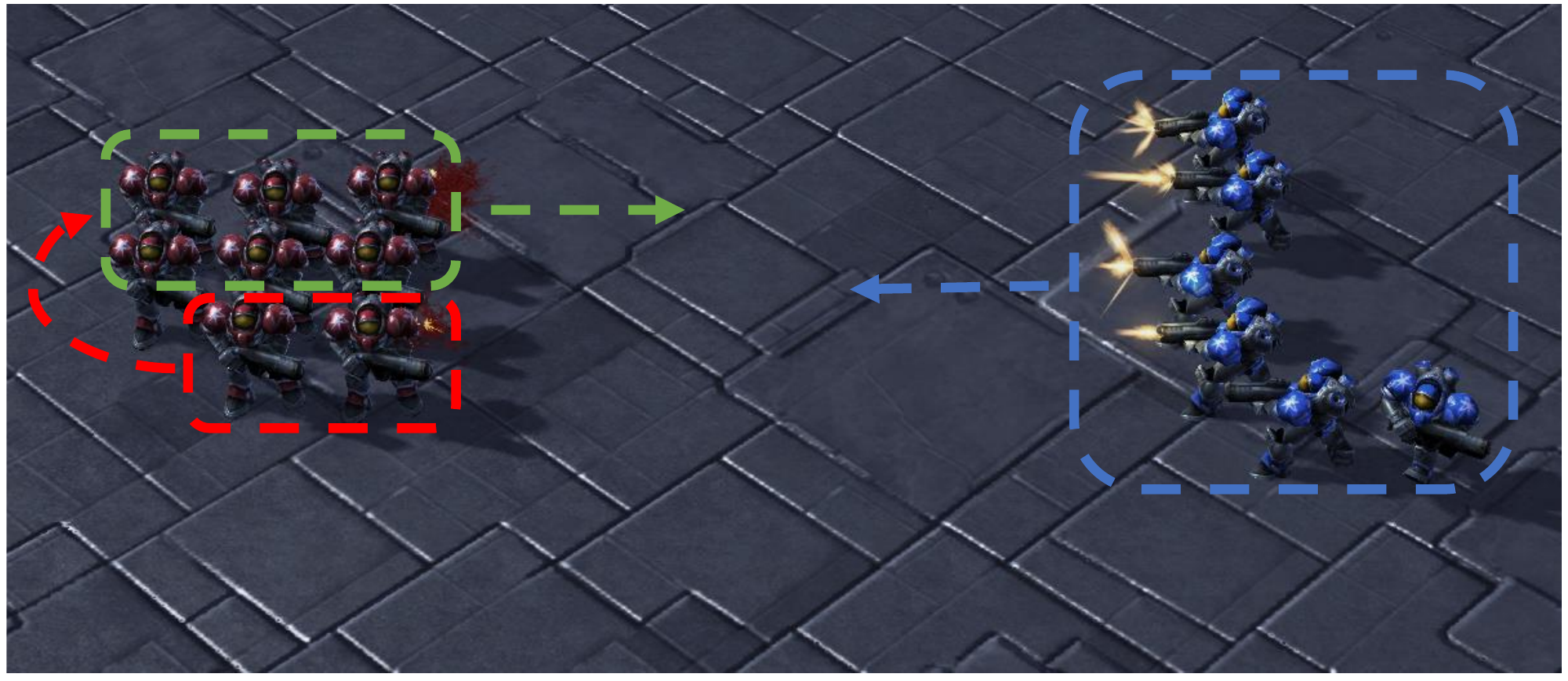}
    \caption{(6+2)m-vs-6m Map in StarCraft II. We customize a map to train the traitors, where the two traitors are circled in red, the six victim agents are circled in green, and the six enemies are circled in blue. The traitors' goal is to reduce the win rate of victim agents.}
    \label{fig:environment}
\end{figure}

\section{Experimental Setup}
\subsection{Environments}
We conduct our experiments on SMAC which is a widely adopted environment for research in the field of cooperative multi-agent reinforcement learning based on Blizzard's StarCraft II RTS game~\cite{samvelyan19smac}. 
SMAC is composed of many combat scenarios with pre-configured maps, where we train the ally units to beat enemy units controlled by the built-in AI with an unknown strategy. 
At each timestep, agents can move or attack any enemies and receive a global reward equal to the total damage done to enemy units.

Different from the original 8m-vs-8m, 5m-vs-6m and other SMAC maps, we have designed two new sets of environments: (6+$N$)m-vs-6m and (8+$N$)m-vs-8m, where $N$ represents the number of traitors and can be 1, 2, or 3. 
As shown in Fig.\ref{fig:environment}, in the original 6m-vs-6m environment, we have inserted 2 traitors whose allegiance is with the victim agents' side. 
The traitors cannot directly attack the victim agents but can disrupt their formation through collisions. 
At each moment, the traitors receive a global reward equal to the negative reward of the victim agents, meaning that the traitors' goal is to minimize the losses of the enemy.

\begin{figure}
    \centering
    \includegraphics[width=0.99\linewidth]{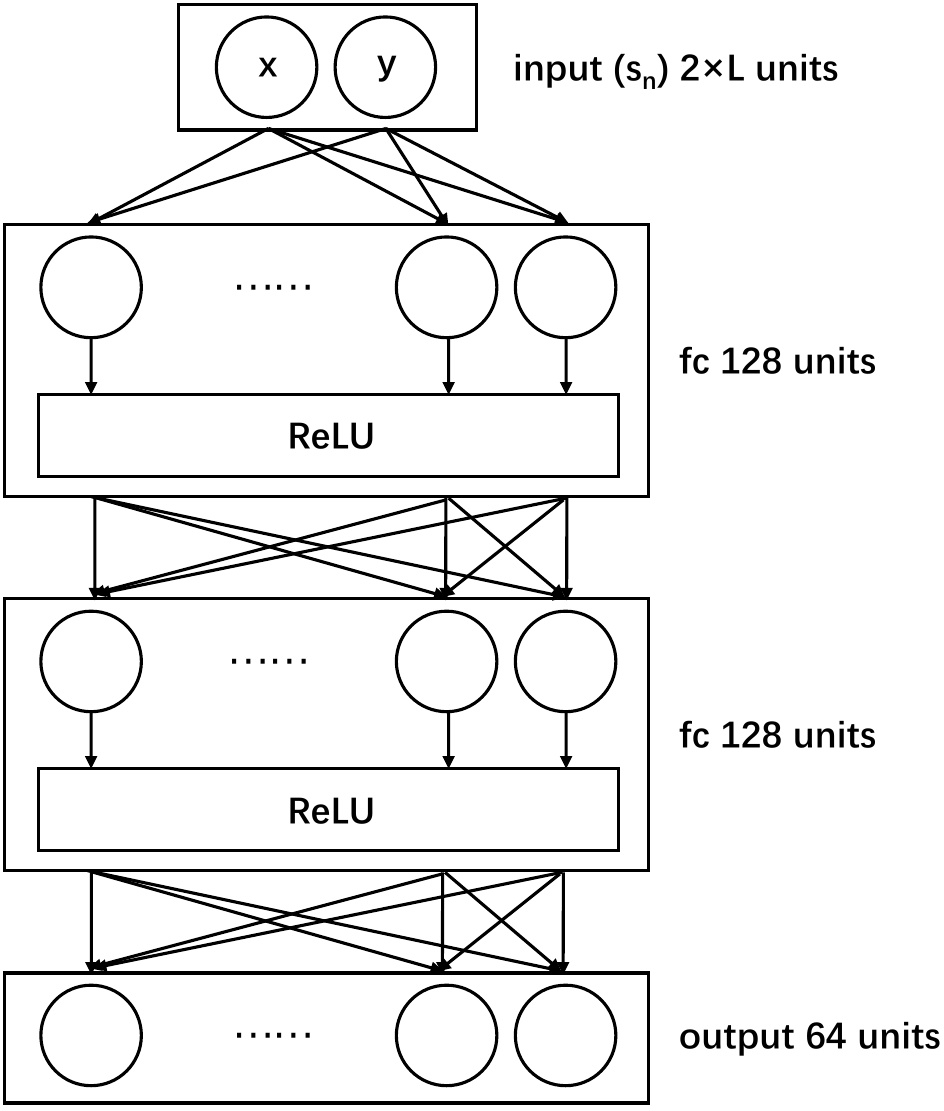}
    \caption{Deep neural network architecture for RND. $N$ is the number of victim agents. This architecture is used in both our method and the ablation experiments.}
    \label{fig:rnd}
\end{figure}

\subsection{MARL Algorithms}
We used three MARL algorithms for our experiments: QMIX, MAPPO and VDN. 
Among these, QMIX and VDN are value-based MARL algorithms, while MAPPO is a policy-based algorithm.
QMIX and VDN are trained for 2,050,000 steps, and MAPPO is trained for 20,050,000 steps.
The network architecture and the hyperparameters of QMIX, MAPPO and VDN are same as that in~\cite{papoudakis2021benchmarking}, which is a benchmark in cooperative MARL tasks. 
In \ref{Different Algorithm}, we compare the results of our method with the baselines under these three different algorithms. 
The algorithm used for all other experiments is QMIX.
\subsection{Baseline Methods}
\label{Baseline Methods}
Under the fixed MARL algorithm of the victim agents, we compared our CuDA2 framework's adversarial attack method with three baselines: \textit{stop}, where traitors remain stationary during the attack; \textit{random}, where traitors perform random actions; and \textit{minus\_r}, where traitors use the same MARL algorithm as the victim agents but with a reward function that is the negative of the victim agents' reward function. 
Our method builds on the \textit{minus\_r} reward function by adding an additional reward through the RND module to incentivize traitors to develop more aggressive attack strategies.
We run each set of experiments five times.

\subsection{RND Architecture}
The Random Network Distillation (RND) architecture is to develop a method that calculates curiosity but that is not attracted by the stochastic elements of an environment. 
The RND module used in this paper is shown in Fig.\ref{fig:rnd}. 
The input $S_n$ is the xy-coordinate of the victim agents, which feeds into the first fully connected (fc) layer consisting of 128 units and utilizing the ReLU activation function to introduce non-linearity. 
This is followed by a second fully connected layer, also with 128 units and ReLU activation, further transforming the data. 
Finally, the processed data is passed through an output layer comprising 64 units, which produces the final output.

\begin{figure*}[htbp]
	\centering
	\begin{minipage}{0.28\linewidth}
		\centering
		\includegraphics[width=0.99\linewidth]{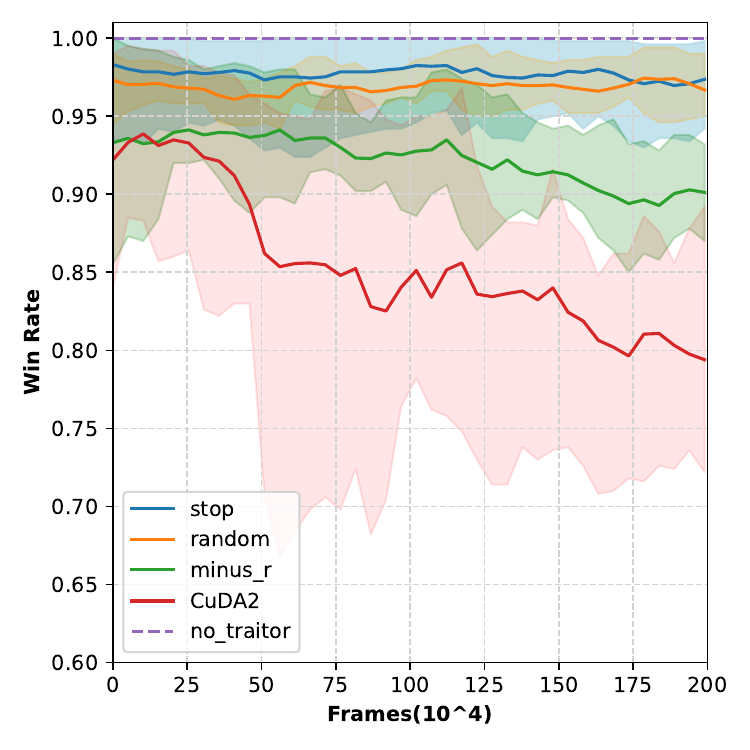} 
            \\\centering(a) VDN
	\end{minipage}
	\begin{minipage}{0.28\linewidth}
		\centering
		\includegraphics[width=0.99\linewidth]{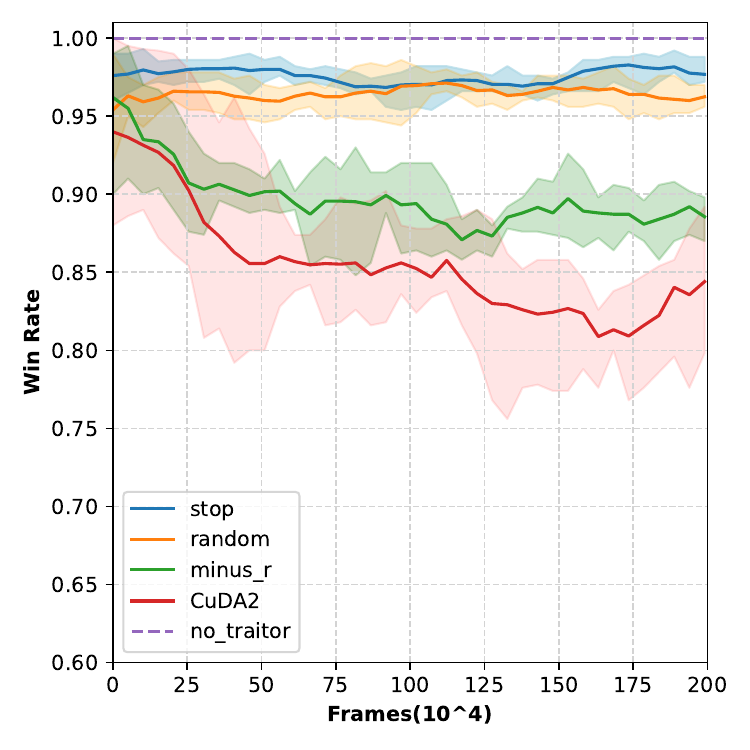} 
            \\\centering(b) QMIX
	\end{minipage}
	\begin{minipage}{0.28\linewidth}
		\centering
		\includegraphics[width=0.99\linewidth]{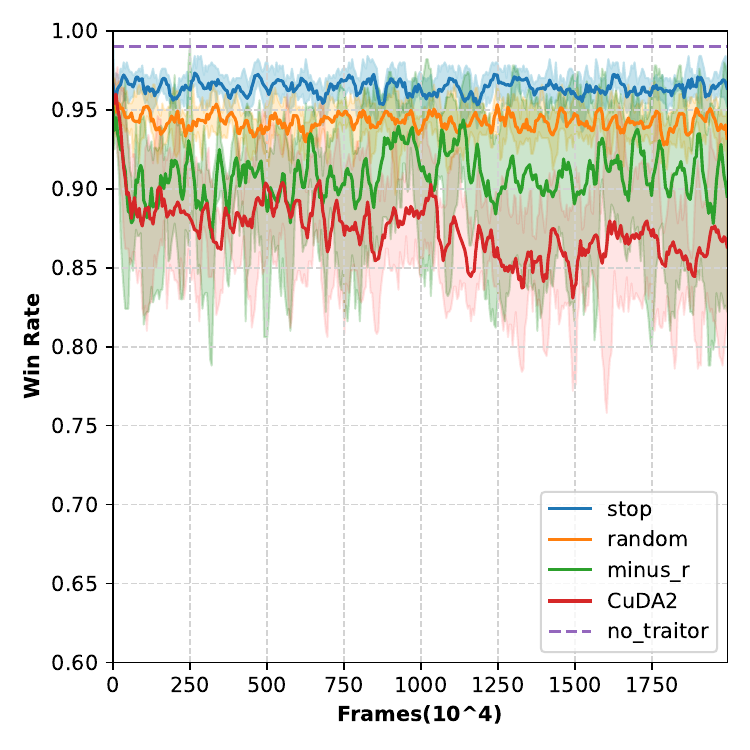}
            \\\centering(c) MAPPO
	\end{minipage}
        \caption{We test our method under different MARL algorithms in (6+1)m-vs-6m maps in comparison to the baseline method.}
        \label{fig:different_algorithms}
\end{figure*}

\begin{figure*}[htbp]
	\centering
	\begin{minipage}{0.48\linewidth}
		\centering
		\includegraphics[width=0.99\linewidth]{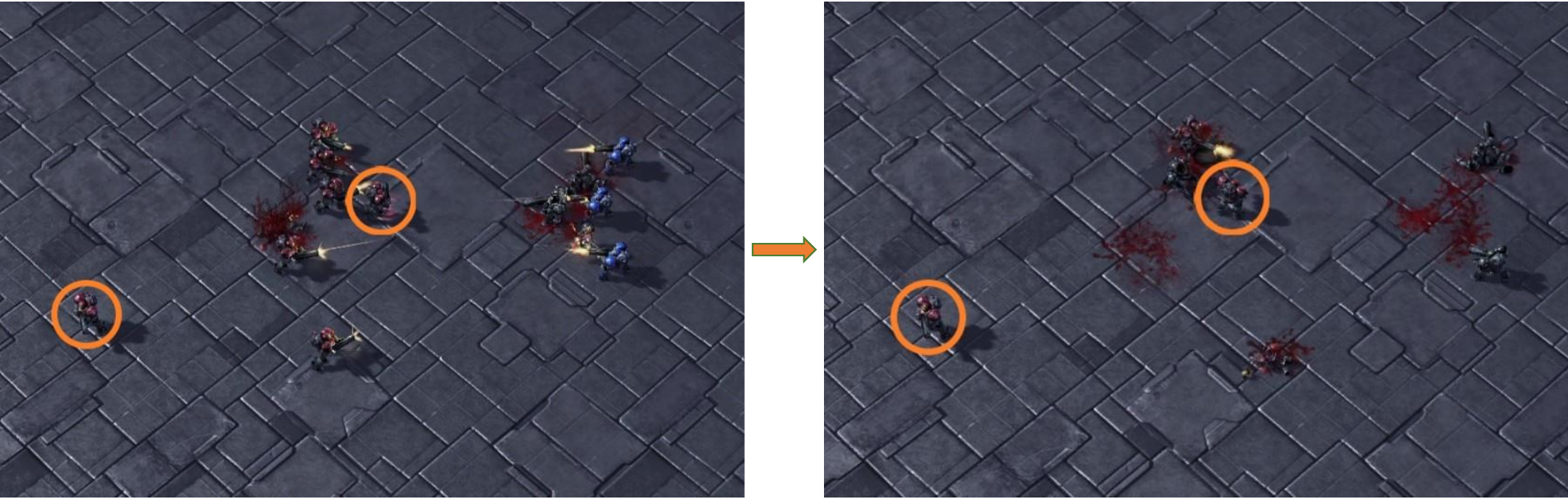} 
            \\\centering(a) stop.
	\end{minipage}
	\begin{minipage}{0.48\linewidth}
		\centering
		\includegraphics[width=0.99\linewidth]{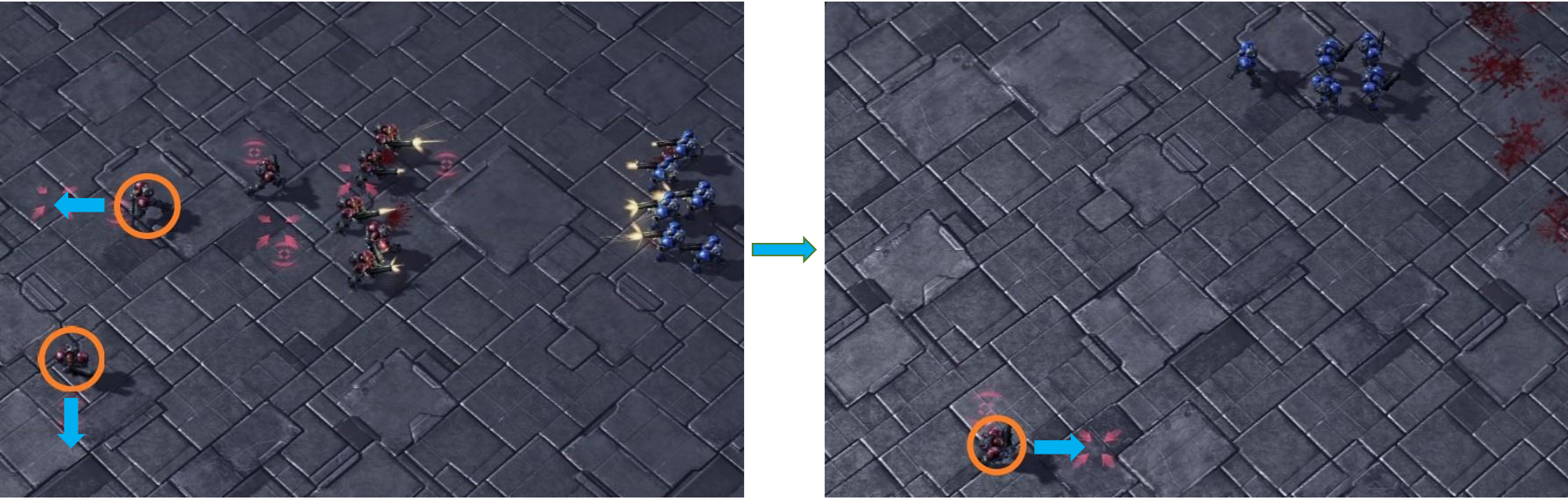} 
            \\\centering(b) random.
	\end{minipage}
    \\
    \begin{minipage}{0.48\linewidth}
		\centering
		\includegraphics[width=0.99\linewidth]{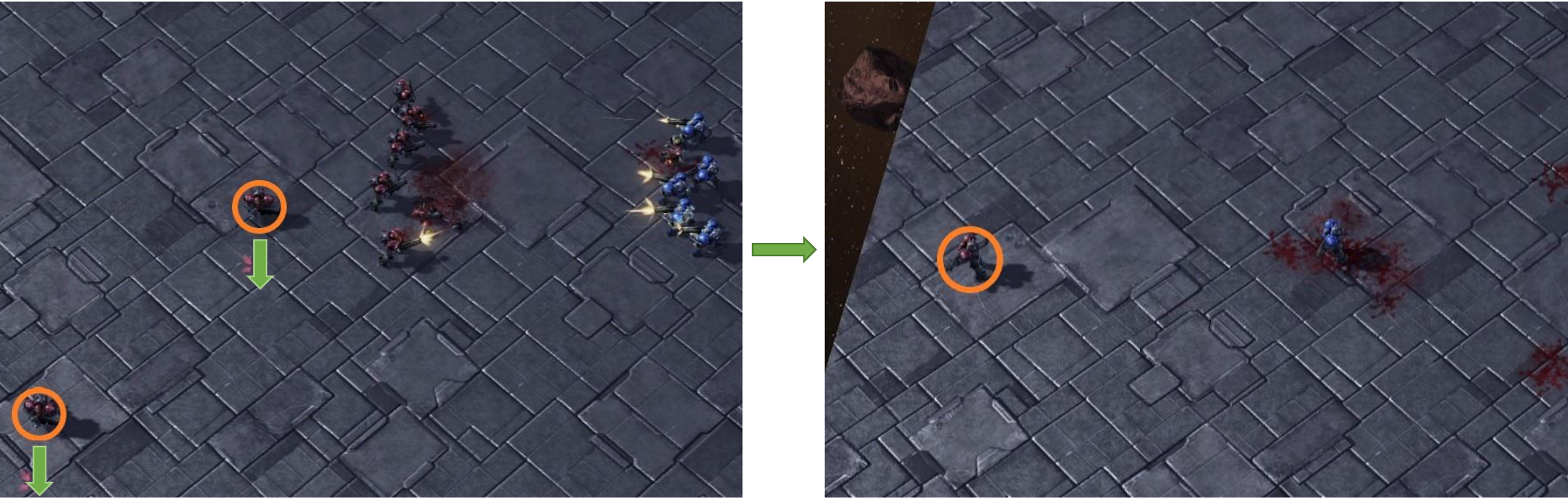} 
            \\\centering(c) minus\_r.
	\end{minipage}
	\begin{minipage}{0.48\linewidth}
		\centering
		\includegraphics[width=0.99\linewidth]{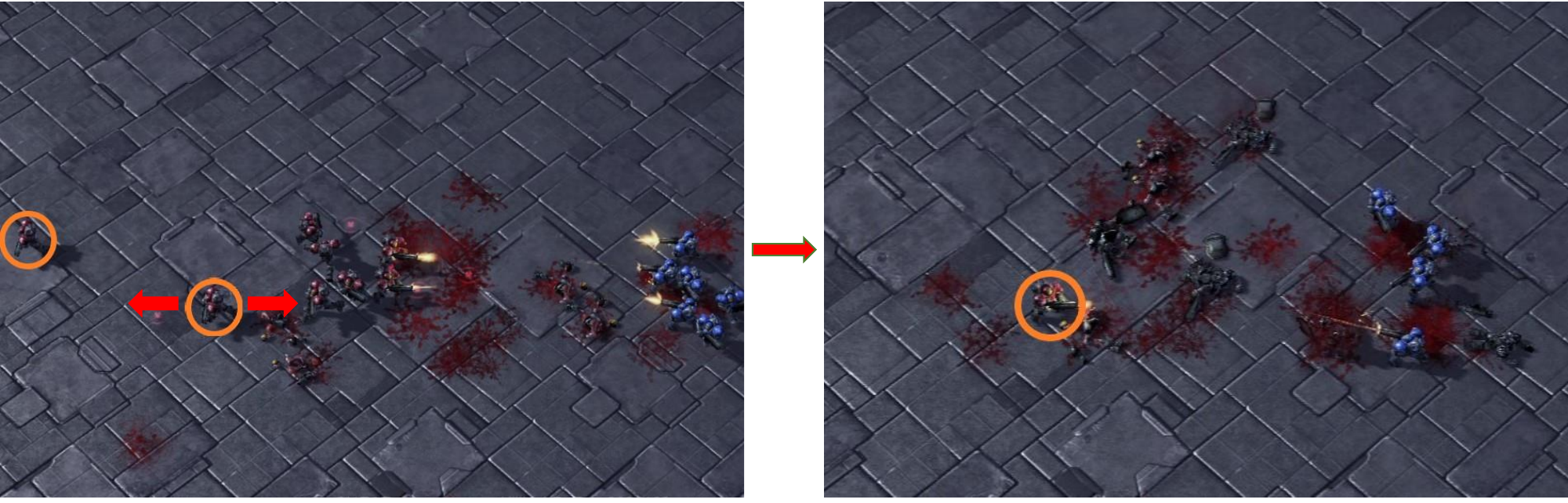} 
            \\\centering(d) CuDA2.
	\end{minipage}
        \caption{Snapshots of our method and baseline method. (a) The traitors remain stationary. (b) The traitors take random actions. (c) The traitors are trained using the same algorithm as the victim agents, with their reward function being the negative of the victim agents' reward. (d) the traitors receive extra rewards provided by the CuDA2 framework.}
        \label{fig:schema}
\end{figure*}

\subsection{Training Details}
Before training the traitors in the CuDA2 framework, we need to pre-train the RND module. 
We perform the pre-training in the baseline where the traitors adopt a \textit{random} action strategy \ref{Baseline Methods}. 
This pre-training offers two benefits: first, it enhances the RND module's sensitivity to unknown states, and second, it reduces the extra reward generated by state changes caused by random actions. 
Additionally, during the training process in the CuDA2 framework, we also update the parameters of the model network within the RND module, allowing it to evolve alongside the traitors’ strategy updates. 
The state obtained from the environment needs to be trimmed to only retain the positional information. 
The trimmed parts include the agents' health and shield information, which gradually decrease over time.
These values range from 0 to $health_{max}$ or $shield_{max}$ in any method, resulting in a consistent distribution across all methods with no variability. 
For the RND module, this information constitutes noise. We need to remove this extraneous information to make the RND module more sensitive to unknown states.

\section{Experiments}
In this section, we first compare the results between our method and baselines. 
Then, we analyze the impact of each module within the CuDA2 framework on the performance of traitor agents, providing additional details and potential insights.
\subsection{Comparison to Baselines}
To validate the effectiveness of our method under different MARL algorithms, we compare the proposed method with the baselines across three MARL algorithms (QMIX, MAPPO, VDN). 
The baselines include \textit{stop}, \textit{random}, and \textit{minus\_r}, as defined in \ref{Baseline Methods}. 
Then, to qualitatively analyze the impact of the number of traitors and the ratio of traitors to allies on the performance of our method and the baselines, we design two experimental environments: 6m-vs-6m and 8m-vs-8m and evaluate the impact on the allies' win rate and the number of allied deaths resulted from adding 1, 2, or 3 traitors.
\subsubsection{Different MARL Algorithms}
\label{Different Algorithm}
As shown in Fig.\ref{fig:different_algorithms}, after introducing a traitor agent into the VDN, QMIX, and MAPPO algorithms, we compared the decrease in the win rates of allies between our method and the baselines. 
The dashed line represents the highest stable win rate that allies could achieve before adding the traitor. 
It can be seen that all three algorithms could achieve nearly 100\% win rates in the 6m-vs-6m environment.
After adding the traitor, our method decreases the win rates of allies to a more apparent degree compared to the baseline methods. 
To be noted, in subsequent experiments where we test the number of traitors, we uniformly used QMIX to train the policy of the victim agents.

\begin{figure*}[htbp]
	\centering
	\begin{minipage}{0.245\linewidth}
		\centering
		\includegraphics[width=0.99\linewidth]{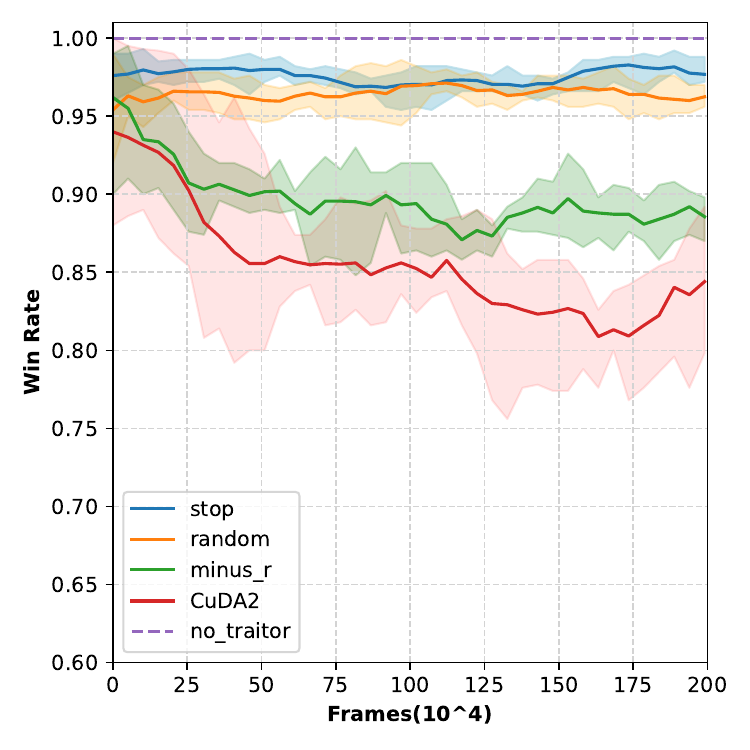} 
            \\\centering(a) (6+1)m-vs-6m.
	\end{minipage}
	\begin{minipage}{0.245\linewidth}
		\centering
		\includegraphics[width=0.99\linewidth]{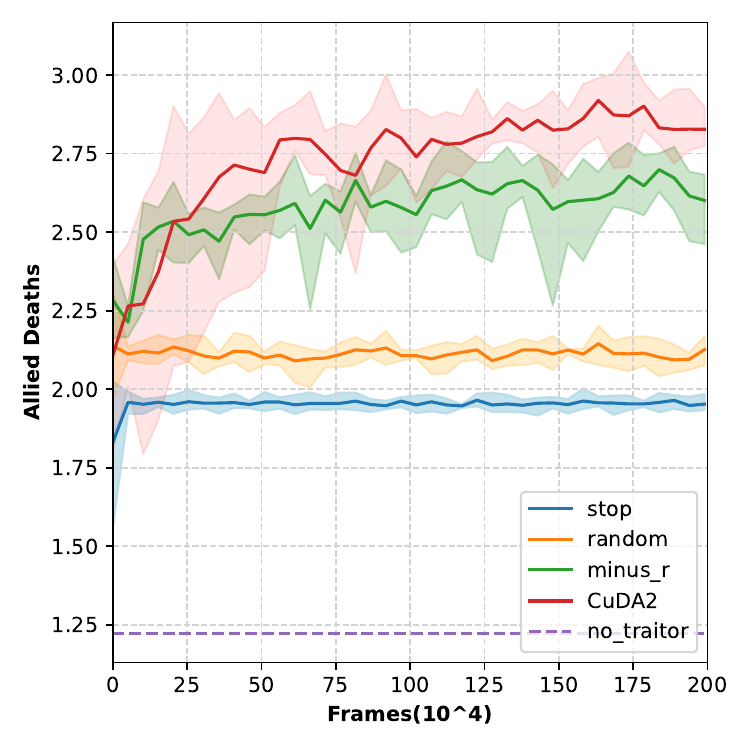} 
            \\\centering(b) (6+1)m-vs-6m.
	\end{minipage}
	\begin{minipage}{0.245\linewidth}
		\centering
		\includegraphics[width=0.99\linewidth]{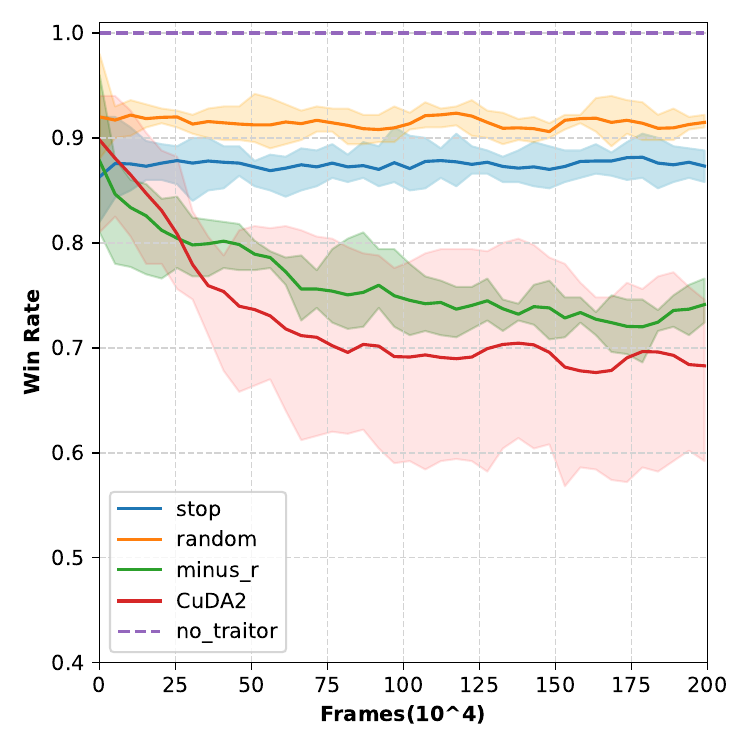}
            \\\centering(c) (8+1)m-vs-8m.
	\end{minipage}
	\begin{minipage}{0.245\linewidth}
		\centering
		\includegraphics[width=0.99\linewidth]{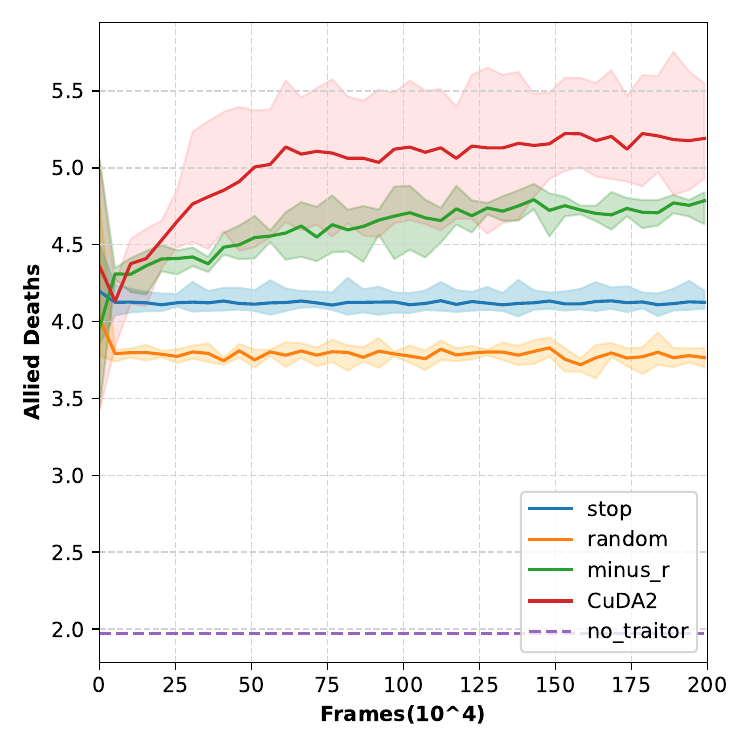}
    		\\\centering(d) (8+1)m-vs-8m.
	\end{minipage}
        \caption{Adding one traitor to the 6m-vs-6m and 8m-vs-8m environments. The number of Allied deaths includes the deaths of traitors. (a) Adding one traitor to the 6m-vs-6m environment, the curve of allied win rate. (b) Adding one traitor to the 6m-vs-6m environment, the curve of allied deaths. (c) Adding one traitor to the 8m-vs-8m environment, the curve of allied win rate. (d) Adding one traitor to the 8m-vs-8m environment, the curve of allied deaths.}
        \label{fig:one_traitor}
\end{figure*}

\begin{figure*}[htbp]
	\centering
	\begin{minipage}{0.245\linewidth}
		\centering
		\includegraphics[width=0.99\linewidth]{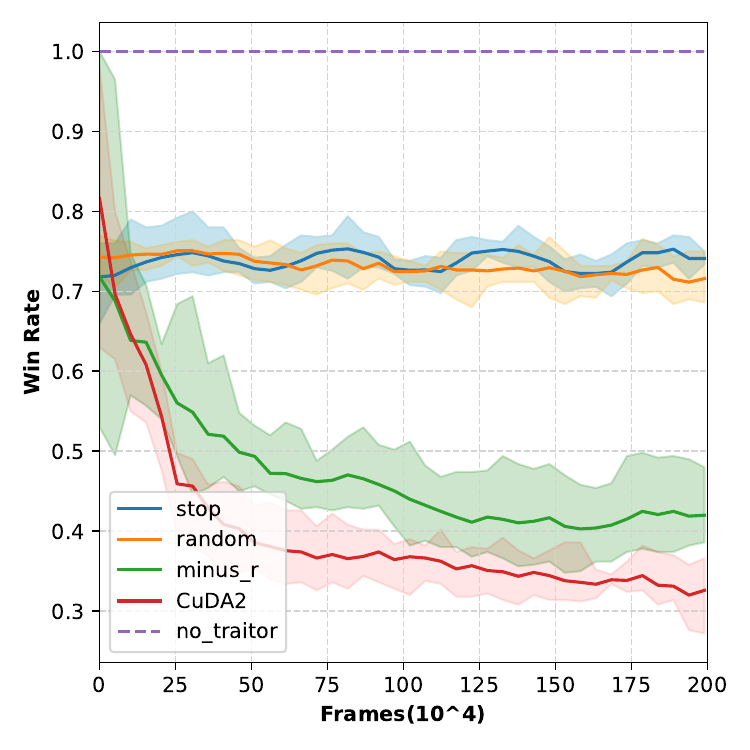} 
            \\\centering(a) (6+2)m-vs-6m.
	\end{minipage}
	\begin{minipage}{0.245\linewidth}
		\centering
		\includegraphics[width=0.99\linewidth]{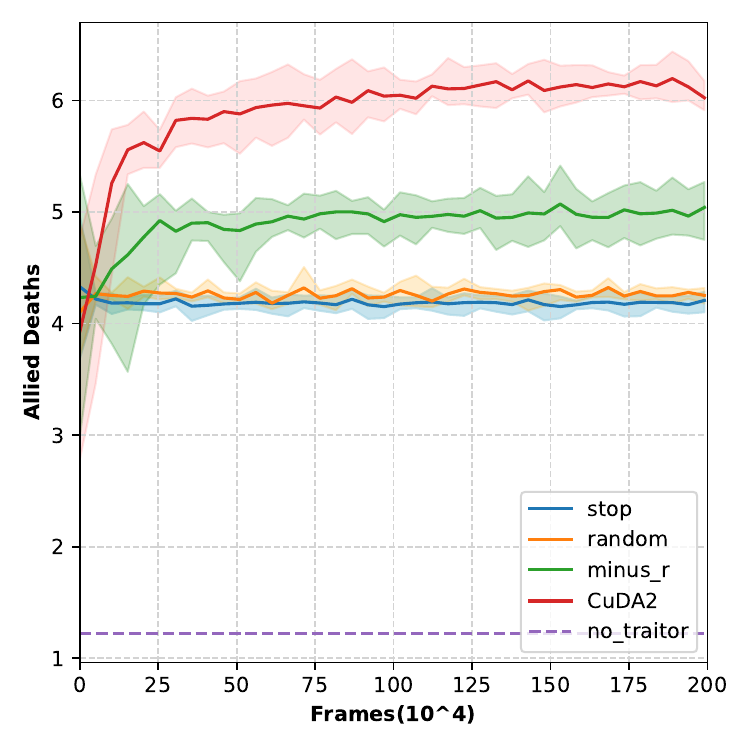} 
            \\\centering(b) (6+2)m-vs-6m.
	\end{minipage}
	\begin{minipage}{0.245\linewidth}
		\centering
		\includegraphics[width=0.99\linewidth]{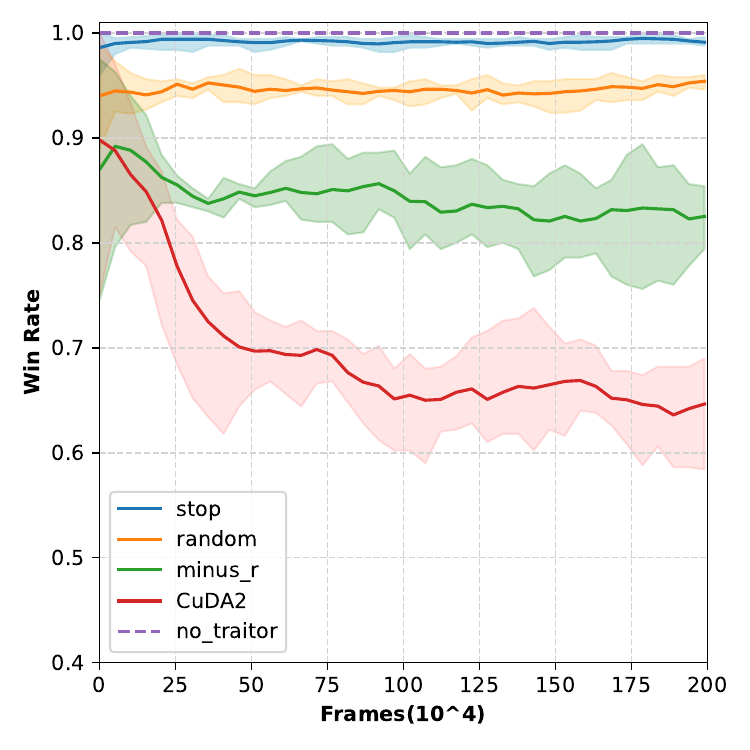}
            \\\centering(c) (8+2)m-vs-8m.
	\end{minipage}
	\begin{minipage}{0.245\linewidth}
		\centering
		\includegraphics[width=0.99\linewidth]{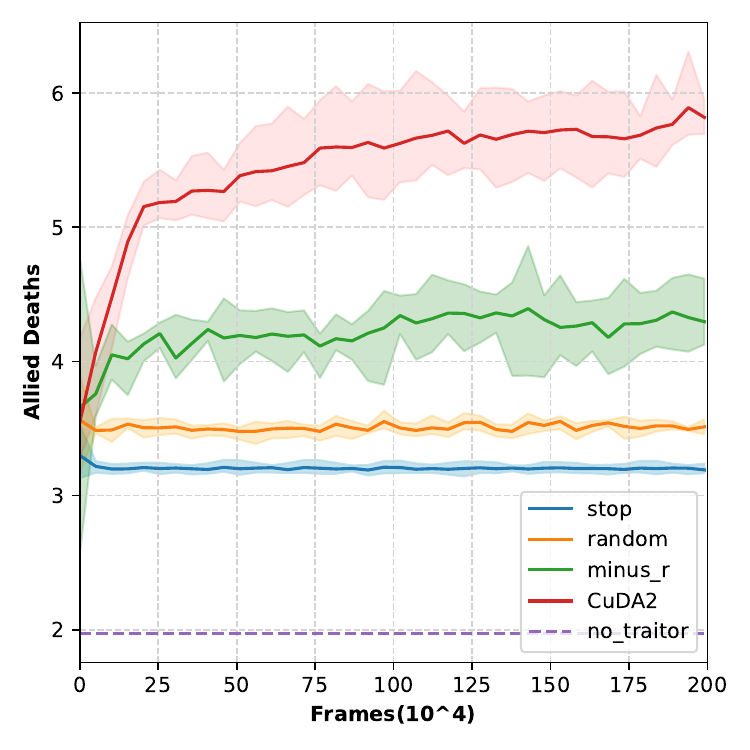}
    		\\\centering(d) (8+2)m-vs-8m.
	\end{minipage}
        \caption{Adding two traitors to the 6m-vs-6m and 8m-vs-8m environments. The number of Allied deaths includes the deaths of traitors. (a) Adding two traitors to the 6m-vs-6m environment, the curve of allied win rate. (b) Adding two traitors to the 6m-vs-6m environment, the curve of allied deaths. (c) Adding two traitors to the 8m-vs-8m environment, the curve of allied win rate. (d) Adding two traitors to the 8m-vs-8m environment, the curve of allied deaths.}
        \label{fig:two_traitors}
\end{figure*}

\begin{figure*}[htbp]
	\centering
	\begin{minipage}{0.245\linewidth}
		\centering
		\includegraphics[width=0.99\linewidth]{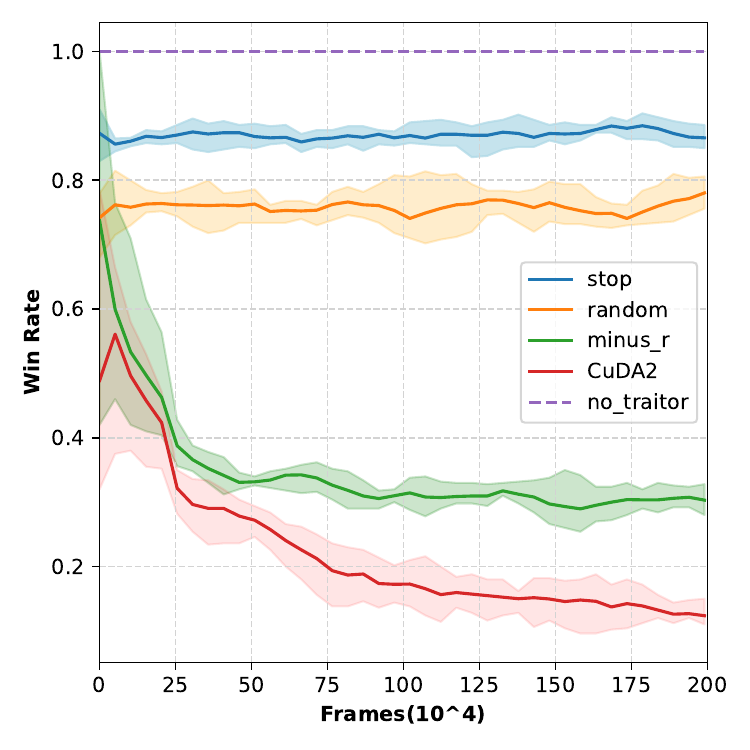} 
            \\\centering(a) (6+3)m-vs-6m.
	\end{minipage}
	\begin{minipage}{0.245\linewidth}
		\centering
		\includegraphics[width=0.99\linewidth]{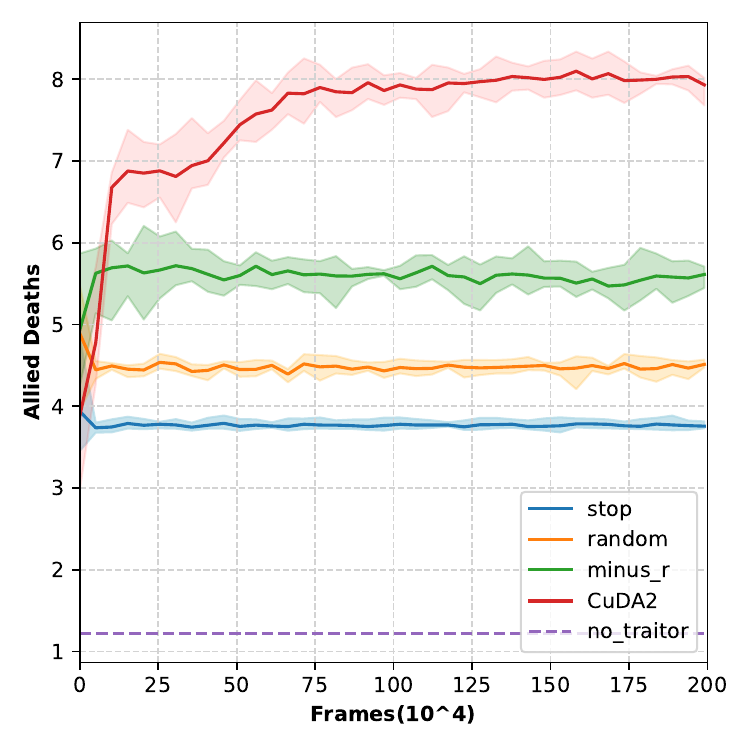} 
            \\\centering(b) (6+3)m-vs-6m.
	\end{minipage}
	\begin{minipage}{0.245\linewidth}
		\centering
		\includegraphics[width=0.99\linewidth]{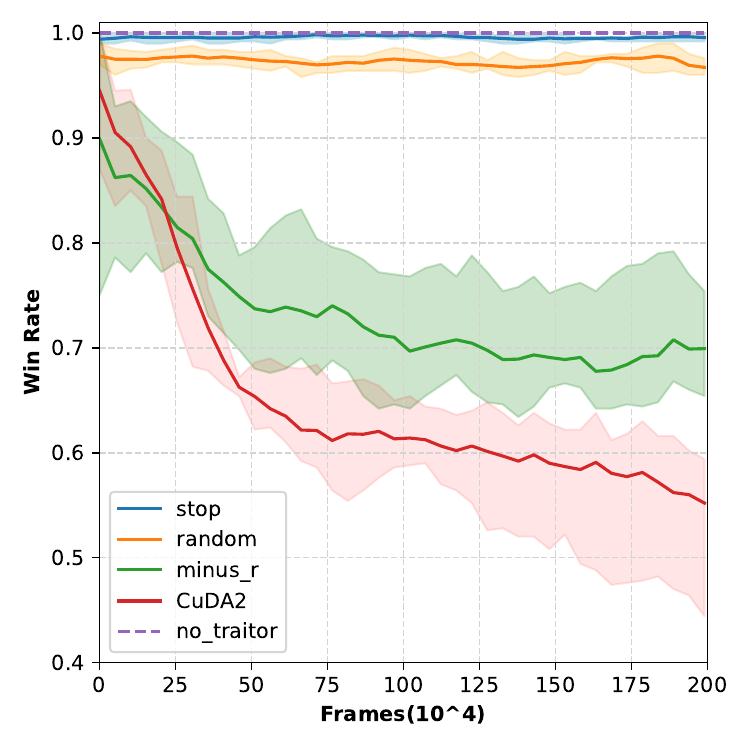}
            \\\centering(c) (8+3)m-vs-8m.
	\end{minipage}
	\begin{minipage}{0.245\linewidth}
		\centering
		\includegraphics[width=0.99\linewidth]{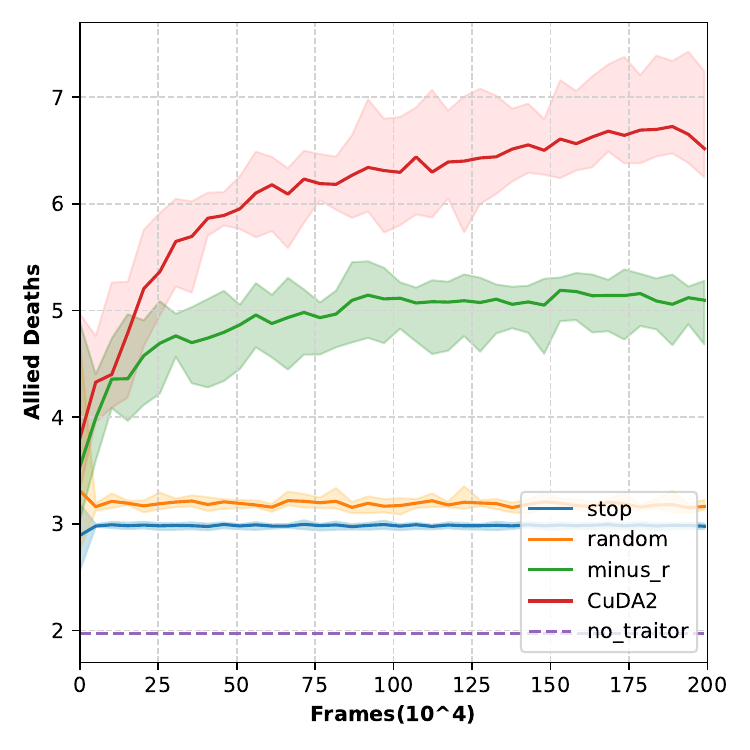}
    		\\\centering(d) (8+3)m-vs-8m.
	\end{minipage}
        \caption{Adding three traitors to the 6m-vs-6m and 8m-vs-8m environments. The number of Allied deaths includes the deaths of traitors. (a) Adding three traitors to the 6m-vs-6m environment, the curve of allied win rate. (b) Adding three traitors to the 6m-vs-6m environment, the curve of allied deaths. (c) Adding three traitors to the 8m-vs-8m environment, the curve of allied win rate. (d) Adding three traitors to the 8m-vs-8m environment, the curve of allied deaths.}
        \label{fig:three_traitors}
\end{figure*}

\begin{figure*}[htbp]
	\centering
	\begin{minipage}{0.245\linewidth}
		\centering
		\includegraphics[width=0.99\linewidth]{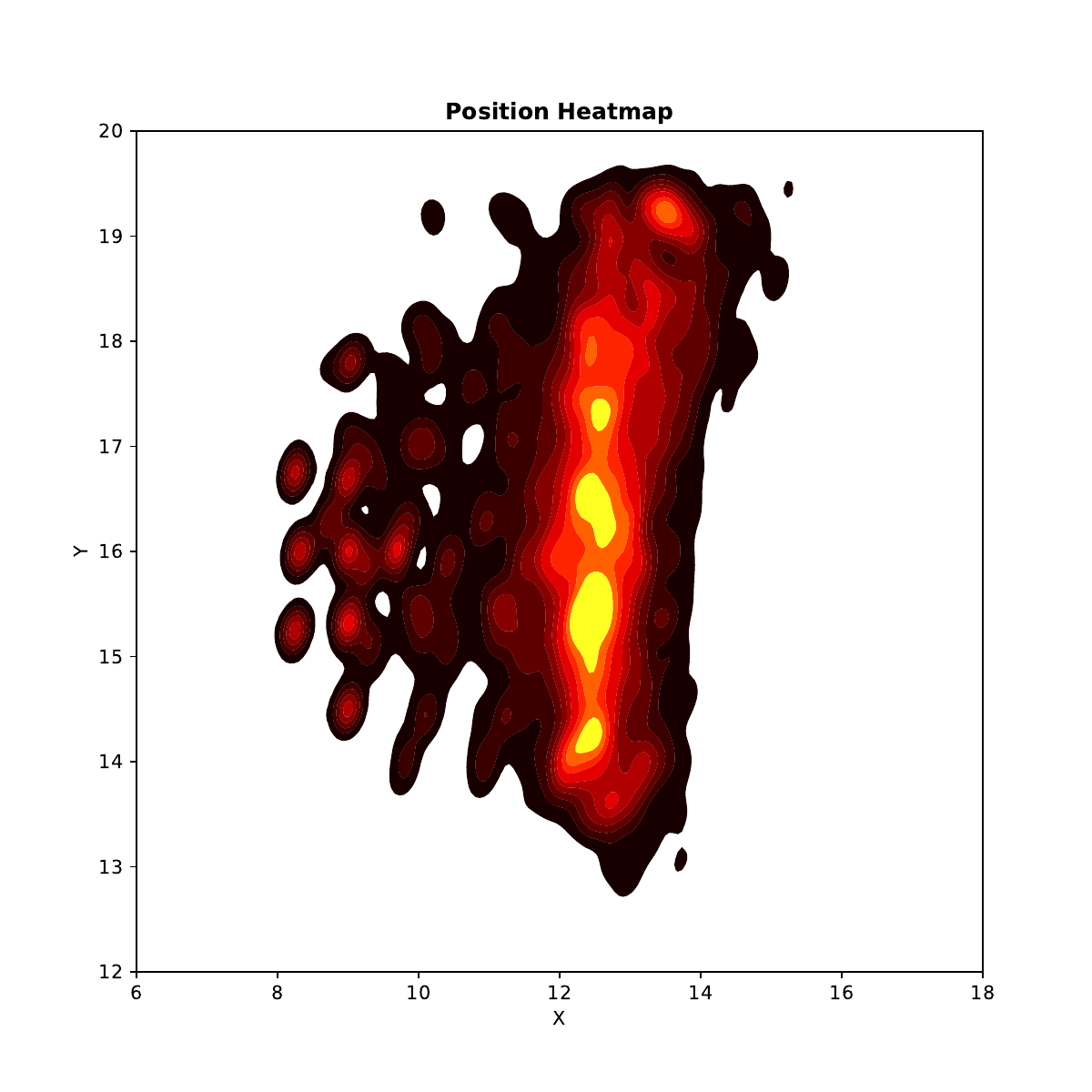} 
            \\\centering(a) Victim agents (RND).
	\end{minipage}
	\begin{minipage}{0.245\linewidth}
		\centering
		\includegraphics[width=0.99\linewidth]{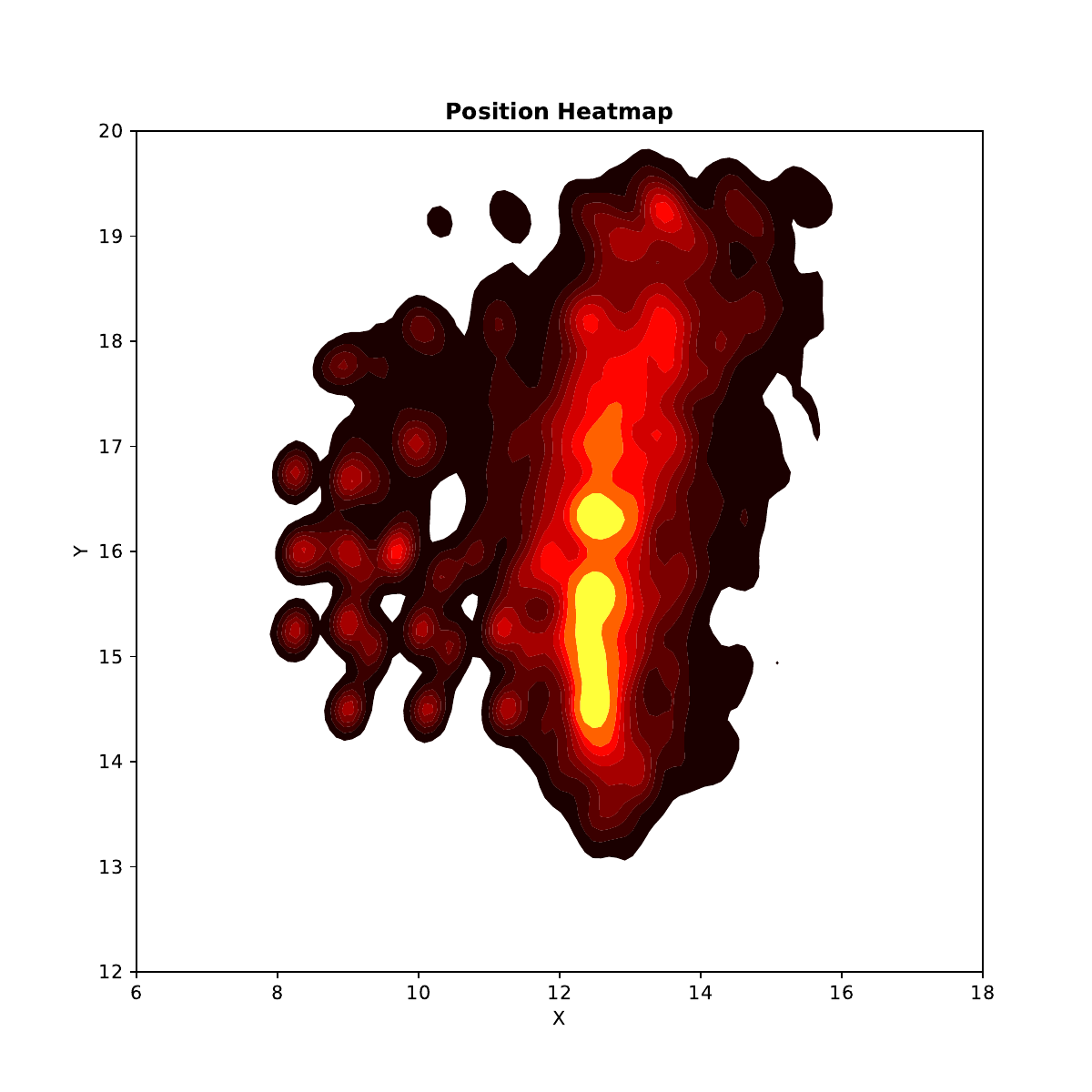} 
            \\\centering(b) Victim agents (CuDA2).
	\end{minipage}
	\begin{minipage}{0.245\linewidth}
		\centering
		\includegraphics[width=0.99\linewidth]{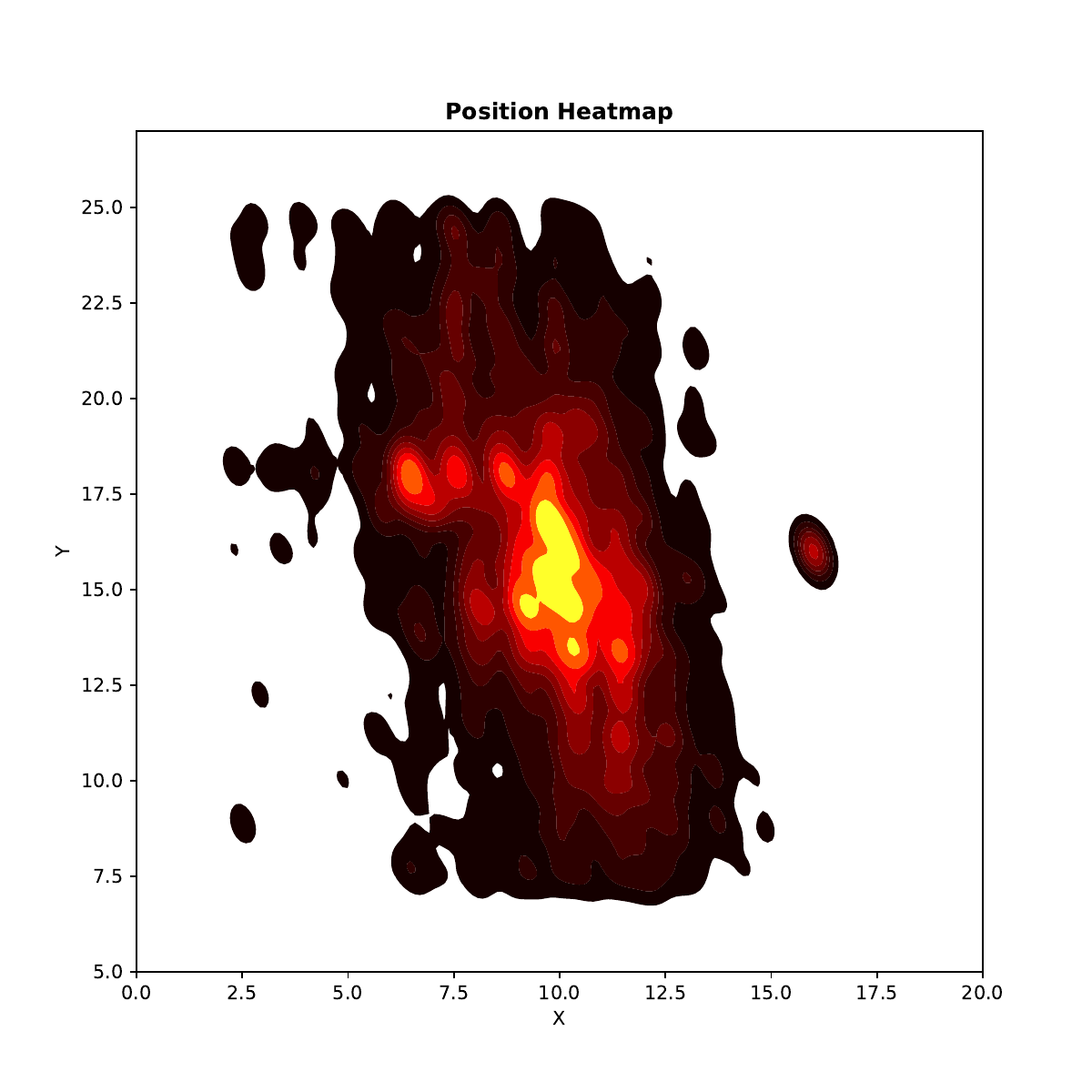}
            \\\centering(c) Traitors (RND).
	\end{minipage}
	\begin{minipage}{0.245\linewidth}
		\centering
		\includegraphics[width=0.99\linewidth]{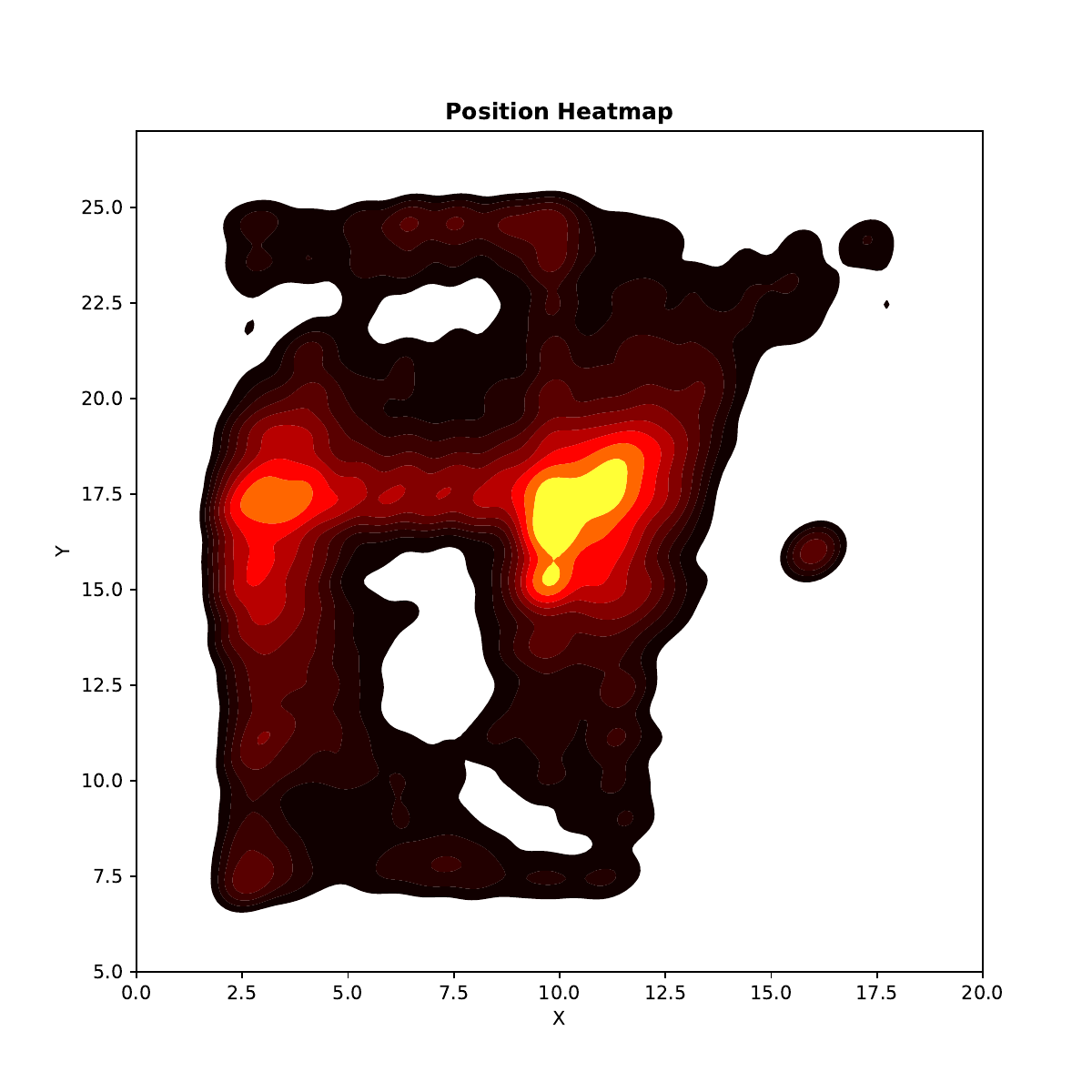}
    		\\\centering(d) Traitors (CuDA2).
	\end{minipage}
        \caption{Position heatmaps of victim agents and traitors under different methods. (a) Position distribution of victim agents with traitors receiving extra rewards from RND. (b) Position distribution of victim agents with traitors receiving extra rewards from CuDA2. (c) Position distribution of traitors with RND providing extra rewards. (d) Position distribution of traitors with CuDA2 providing extra rewards.}
        \label{fig:heatmap}
\end{figure*}

\subsubsection{Number of Traitors}
In this section, we compare the impact of introducing different numbers of traitors on the win rates and death counts of the victim agents. 
We set up two environments: 6m-vs-6m and 8m-vs-8m, and add 1, 2, or 3 traitors to these environments. 
In Fig.\ref{fig:schema}, we show the behavior of two traitors in the 6m-vs-6m environment using CuDA2 and the baseline methods. 
In Fig.\ref{fig:schema}(a), the traitors remain stationary throughout the process. 
In Fig.\ref{fig:schema}(b), the traitors take random actions and one of them moves to the edge of map due to random movements, causing the enemies to be unable to detect and kill it. 
In Fig.\ref{fig:schema}(c), traitors trained with the minus reward function of victim agents exhibit runaway behavior. 
In Fig.\ref{fig:schema}(d), the traitors trained with CuDA2 collide with the victim agents at the initial stage of each episode and then move back and forth at the edge of the victim agents' observation range, thereby influencing the victim agents' decision-making.

As shown in Fig.\ref{fig:one_traitor}, Fig.\ref{fig:two_traitors} and Fig.\ref{fig:three_traitors}, the win rates of the allies decreases more significantly as the number of traitors increases. 
Especially when the number of traitors is two or more, our method exhibits more significant impact on the allies' win rate and death count compared to the \textit{minus\_r} baseline. 
We assume that this is because traitors in CuDA2 framework disrupt the formation of victim agents through collisions, causing state-action pairs that victim agents have never encountered during previous training, rendering their original strategies ineffective and leading to their defeat by the enemies one by one.
Except for the scenario with a single traitor, the attack effect is more pronounced in the 6m-vs-6m environment compared to the 8m-vs-8m environment when the same number of traitors is added, due to the higher ratio of traitors to victim agents in the 6m-vs-6m environment.


\subsection{Ablation Study}
In the 8m-vs-8m environment with two traitors, we evaluate the results of using RND alone to give traitors intrinsic rewards and compared them with our method. 
To be specific, our CuDA2 framework incorporates the dynamic PBRS method to help traitor agents learn optimal attack policy compared to just using RND.
We analyze the behavioral differences between the traitors and victim agents by plotting heat maps of their positions. 
In the experimental scenario, the victim agents tend to start from the left side of the map and then spread out in a line in the middle to attack the enemies.
As shown in Fig.\ref{fig:heatmap}(a) and (b), the movement range of victim agents under RND is smaller compared to CuDA2, indicating that RND technique results in less effect on the victim agents' policy. 
In Fig.\ref{fig:heatmap}(c) and (d), there is a significant difference in the behavior of traitors under RND and CuDA2. 
RND traitors tend to move near the central area of the map while CuDA2 traitors choose to move away from the combat area to avoid their own death after disrupting the victim agents in the center. 
Therefore, in terms of both influencing the victim agents' policy and ensuring the traitors' survival, CuDA2 performs better than RND.

\section{Conclusion and Future Work}
This paper addresses the challenge of incorporating traitor agents into Cooperative Multi-Agent Reinforcement Learning (CMARL).
Unlike previous research about adversarial attacks, the traitors in our setting cannot directly interfere with the states and actions of the victim agents.
Instead, we control the behavior of the traitors to indirectly influence the formation and positions of the victim agents.
We formalize this problem as a Traitor Markov Decision Process (TMDP), where the traitor agents aim to minimize the cumulative discounted reward of the victim agents when the policies of the victim agents are fixed.
We then introduce the Curiosity-Driven Adversarial Attack (CuDA2) framework.
By pre-training the Random Network Distillation (RND) module and shaping intrinsic rewards using the dynamic Potential-Based Reward Shaping (PBRS) method, CuDA2 ensures the invariance of the traitors' optimal policy while guiding the traitors to perform more aggressive attacks.
Experimental results from comparisons with baselines and ablation studies validate the effectiveness of the CuDA2 framework from multiple perspectives.
As our method is an efficient curiosity-driven adversarial attack algorithm, we will focus on detecting the presence of traitors and defending against their attacks in the future works.

\bibliographystyle{IEEEtran}
\bibliography{template}

\end{document}